\documentclass[sigconf]{acmart}
\renewcommand\footnotetextcopyrightpermission[1]{} 
\usepackage{xcolor}

\usepackage{multirow}
\usepackage{pgfplots}
\pgfplotsset{width=10cm,compat=1.9}

\usepackage{graphicx}
\usepackage{pdfpages}

\AtBeginDocument{%
  \providecommand\BibTeX{{%
    \normalfont B\kern-0.5em{\scshape i\kern-0.25em b}\kern-0.8em\TeX}}}

\begin{document}

\title{}




\includepdf{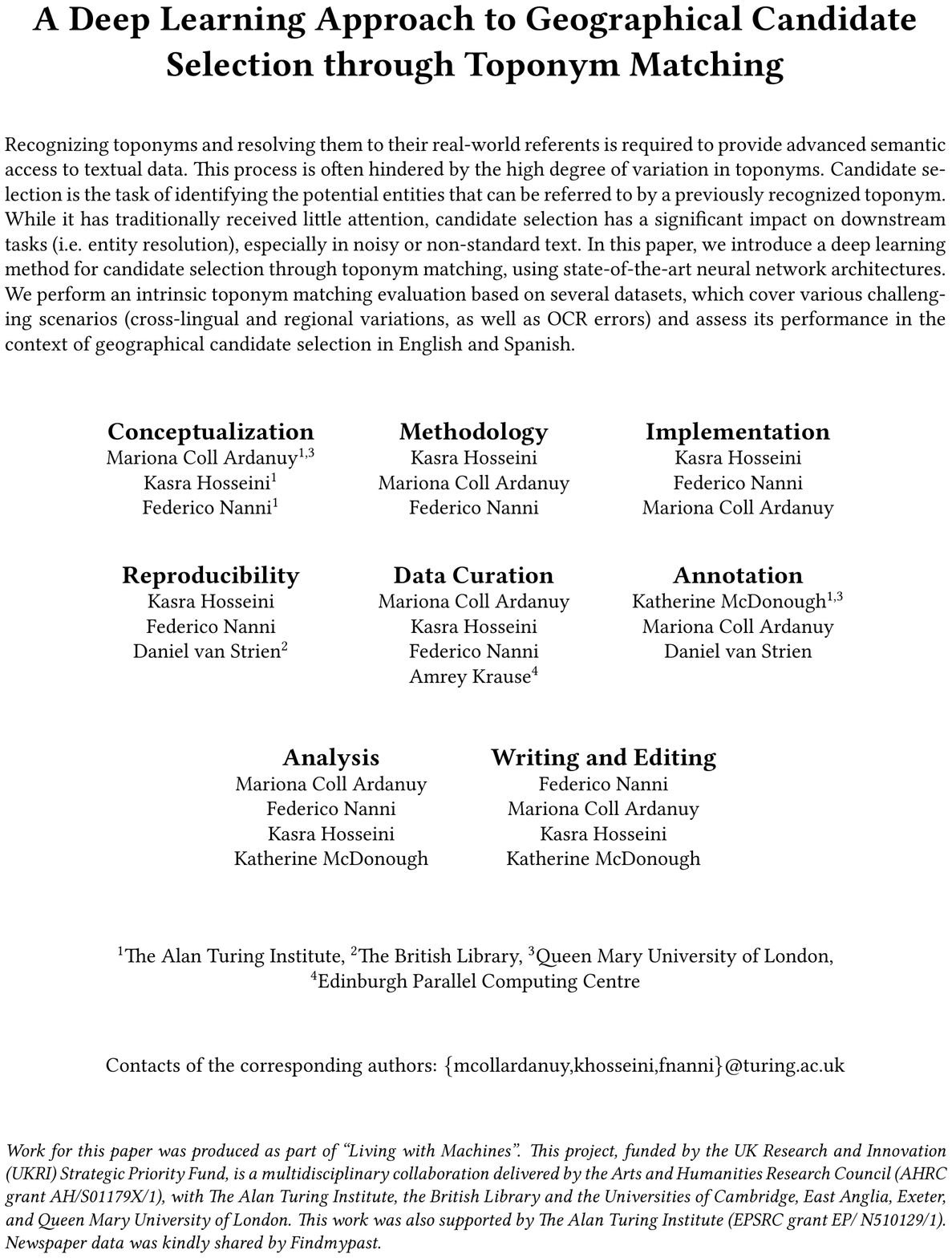}

\newpage

\maketitle
\section{Introduction}
With increasingly larger amounts of unstructured text becoming digitally available in many different fields, the need for robust geographically-aware retrieval of information from large textual collections is now more urgent than ever. 
Textual data is often deeply geographical and it has been shown that geographic queries make for a large part of all search queries \cite{aloteibi2014analyzing,gan2008analysis,gregory_geoparsing_2015,purves_geographic_2018}. Toponym resolution is a class of entity linking that focuses specifically on geographical entities. Given a toponym (i.e.~a geographical name) that has been recognized in text,\footnote{We do not consider toponym detection as part of the toponym resolution task in this paper. There is a large body of research in the natural language processing community that deals with the specific problem of named entity recognition, of which toponym detection is a part.} its aim is to resolve it to its spatial footprint (often represented as a set of coordinates that define its location on the Earth's surface). This step requires an external source of knowledge which usually comes in the shape of a gazetteer, that is, a dictionary of geographical entities with their associated alternative place names and geospatial information. On the other hand, candidate selection is the task of identifying the potential entities that can be referred to by a named entity recognized in text. As the intermediary step between named entity recognition and the downstream task of entity disambiguation, candidate selection is an integral part of entity linking. And yet, it has often been an overlooked component of the entity linking pipeline, even though it has been shown to have a significant impact on the final performance \cite{hachey2013evaluating,quercini2010determining}, especially in noisy or non-standard text.

Toponyms are particularly prone to name variations and changes, which can arise from multiple causes, such as regional spelling differences, diachronic spelling variation, and change of the geopolitical status \cite{butler_alts_2017}. In toponyms, variation is common not only at a token-level (e.g.~`Republic of Ireland' matching `Ireland'), but also at a character-level (e.g.~`Killarra' for `Killala'), and at both token- and character-level (e.g.~`Canouan' and `Cannouan Island'). In addition to these, noisy text often presents other types of character-level variations, such as spelling errors, typographical errors, and OCR errors (e.g.~`Worchestershire' for `Worcestershire', or `Cockcnnotith' for `Cockermouth'). The number of potential variations can be very high, and yet candidate selection should ensure that the correct location is provided among the pool of retrieved entities.

In this paper, we present a new and flexible deep learning approach to geographical candidate selection through toponym matching, which is specifically tailored to dealing with these challenges characteristic of noisy scenarios. Our method consists of two main components: \textbf{(1)} toponym matching, formulated as a binary classification of toponym query-candidate pairs, and \textbf{(2)} candidate selection, formulated as a ranking task where the aim is to rank the good candidates first while minimizing the presence of noisy candidates. The main contributions of this paper are:

\begin{itemize}
    \item A new flexible, user-friendly, efficient software library, with extensive documentation for performing candidate selection through fuzzy string matching. We discuss its relevance and application in the context of geographical candidate selection, and evaluate its performance and efficiency on datasets of various sizes. We call our method DeezyMatch (\emph{DEEp fuzZY string MATCHing}).
    \item New realistic datasets for the evaluation of the toponym matching methods. These datasets cover a wide range of challenging scenarios (e.g.~cross-lingual, diachronic, and regional variations, as well as OCR errors).
    \item A comprehensive evaluation framework for the task of geographical candidate selection in the downstream task of toponym resolution on noisy or non-standard datasets (i.e.~two existing datasets in English and Spanish and one new manually annotated dataset of English nineteenth-century OCR'd text). We conduct an extensive quantitative evaluation covering both the binary classification of toponyms in these datasets and the ranking of potential candidates in real toponym resolution scenarios with models created from these datasets.
\end{itemize}

Our method has been designed to be as language-independent as possible. It only relies upon a character tokenizer when processing the string inputs and a reference gazetteer. We have tested its downstream application on datasets from different languages, time periods and origins, from seventeenth century Latin America to nineteenth century Britain and the United States. All codes, datasets, gazetteers and evaluation settings are openly available to support research reproducibility and to foster the use of \textit{DeezyMatch} in other downstream tasks.\footnote{DeezyMatch codes can be found here: \url{https://github.com/Living-with-machines/DeezyMatch/}. For a more detailed description of the DeezyMatch architecture and functionalities, see \citet{hosseini2020deezy}. All experiments can be found here: \url{https://github.com/Living-with-machines/LwM_SIGSPATIAL2020_ToponymMatching}. We provide all resources to allow full reproducibility of the results.}

\section{Related Work}\label{sec:related_work}
To date, most entity linking and toponym resolution systems have approached candidate selection by performing exact or partial string matching between the mention of the toponym in a text and a name variation of the entry in a knowledge base (KB) for a specific entity (e.g. `NYC' for `New York'). Well-established entity linking pipelines such as the ones presented by \citet{ferragina2010tagme,mendes2011dbpedia,raiman2018deeptype,sil2018neural} and \citet{moro2014entity} depend on exact or super-string matching to select the set of potential candidates a query can refer to. This approach to select candidates relies on the assumption that the mention is present as the name variation of a specific entity in the KB. There have been a number of studies on enriching the entries in a KB with alternate names (such as abbreviations, historical names, or names in other languages) \cite{bunescu2006using,cucerzan2007large,moro2014entity}. Thanks to such studies, most of the effort of the research community invested in candidate selection has been on developing algorithms for scoring and ranking a set of retrieved candidates \cite{ganea-hofmann-2017-deep,le-titov-2019-boosting,martins-etal-2019-joint}, while less effort has been put into dealing with candidate selection in noisy text. Nevertheless, even a KB highly enriched with name variations will not cover all possible name variations (especially of less popular entities), or spelling mistakes and OCR errors.

Alternatives to perfect match linking include the adoption of edit-distance techniques, such as Levenshtein distance \cite{mcnamee2011cross,moreno2017combining}, but these methods suffer from poor scalability. More recently, researchers have proposed deep learning solutions to address this problem. \citet{le-titov-2019-distant} use a noise detector in their entity linking system that operates at the token level (e.g.~`Bill Clinton (President)' matching `Presidency of Bill Clinton'), which learns true matchings from lists of positive and negative candidate pairs. \citet{tam2019optimal} have recently presented STANCE, a model for computing the similarity between two strings by encoding the characters of each of them, aligning the encodings using Sinkhorn Iteration, and scoring the alignments using a convolutional neural network.

The most similar work to ours is by \citet{santos2018toponym}, who proposed a deep learning architecture using Gated Recurrent Units (GRUs) to classify pairs of toponyms as either potentially referring to the same entity or not. The method is trained and intrinsically evaluated on a large dataset collected from the GeoNames gazetteer.\footnote{\url{http://www.geonames.org}.} This dataset is composed of 5 million positive and negative toponym pairs. Our work builds on this by leveraging on current research in natural language processing (NLP), and expands it in several different directions: by supporting various state-of-the-art neural network architectures, allowing the application of an existing model to new data, offering the possibility of further fine-tuning it, and employing it for the task of candidate ranking. More critically, our approach allows the candidate ranking component to be seamlessly integrated into entity linking and toponym resolution pipelines. In the remainder of the paper, we describe our method and assess its performance in the context of geographical candidate selection.

\section{Method}
In this section, we provide a brief overview of DeezyMatch, a free, open-source library written in Python for fuzzy string matching and candidate ranking, and show how it can be used for the task of geographical candidate selection. DeezyMatch consists of two main components: a pair classifier, which we use for the subtask of \textit{toponym matching} and is described in section \ref{subsec:method_topmatching}, and a candidate ranker, which is used for the task of \textit{geographical candidate selection} and is described in section \ref{subsec:method_candselection}.

    \subsection{Toponym Matching}\label{subsec:method_topmatching}
    The DeezyMatch pair classifier component is largely inspired on previous work by \citet{santos2018toponym} for toponym matching, which is formulated as a binary classification task of toponym query-candidate pairs. The authors developed a siamese deep neural network for binary classification of toponym pairs implemented using \texttt{Keras} \cite{chollet2015keras}.
    DeezyMatch builds upon the neural network architecture of \citet{santos2018toponym} and extends it to allow more control on its architecture and each of its components and parameters, direct application on unseen data, and further fine-tuning of already trained models. DeezyMatch is designed following a modular approach, and has been implemented in PyTorch and tested on both CPU and GPU. It is not fixed to a pre-defined architecture: the user can specify the preferred architecture (GRU, LSTM, or RNN); moreover, the dimensionality of the hidden units in the recurrent neural networks, fully connected layers and embeddings can be changed in the input file. The user can choose to work with a forward or bi-directional RNN/GRU/LSTM architecture and can specify the number of layers in the networks. The preprocessing steps and other hyperparameters, such as learning rate, number of epochs, batch size, maximum sequence length and dropout, can all be as well changed in the input file.
    
    During training, a dataset of string pairs is read, preprocessed, and strings are converted into dense vectors. Instead of having a fixed two-fold cross-validation as in \citet{santos2018toponym}, we split the dataset into training, validation and test sets (the ratio of which is specified by the user).\footnote{DeezyMatch supports character-, word-, and ngram- tokenization. Preprocessing steps include lower-casing, stripping, dealing with missing characters in the vocabulary (particularly in the case of fine-tuning) and normalizing strings.} The resulting model can be further fine-tuned by other datasets, an approach especially promising where only limited training examples are available.\footnote{Experiments on the impact of fine-tuning on both toponym matching and candidate selection are underway.} In contrast to \citet{santos2018toponym}, DeezyMatch provides functionality for model inference where a trained model can be applied to other datasets, not used in training and validation, and evaluated through various metrics (loss and precision/recall/F1 scores).
    
    Section~\ref{subsec:deezy_models} summarizes the model architectures and the choice of hyperparameters used in this study.
    
    
    \subsection{Candidate Selection}\label{subsec:method_candselection}
    The \textit{pair classifier} component described in section \ref{subsec:method_topmatching} returns a classifier trained to capture the transformations present in the input dataset of toponym pairs. The \textit{candidate selection} component then uses this model to retrieve potential candidates for a given toponym query --- or set of queries --- from a KB. This is achieved through the following steps:
    \begin{enumerate}
    
        \item[(0)] \textbf{Generating gazetteer vector representations:} a trained DeezyMatch model is used in this initial step to generate vector representations for all the alternate names (i.e.~the set of all locations' toponyms) in the KB or gazetteer to which we want to link our toponym queries. This step is done only once for each model and gazetteer. The vectors (e.g.~forward/backward vectors in a bi-directional neural network) are then combined to form one file containing all the gazetteer vectors.
    
        \item[(1)] \textbf{Generating query vector representations:} the same mo-del used to generate gazetteer vector representations in the previous step reads in a set of toponym queries (e.g.~toponyms recognized in a text) and generates a vector representation for each query term. As above, the vectors are then combined to form one file containing all the query vectors.
         
        \item[(2)] \textbf{Ranking candidates:} The representations generated in the previous steps encode toponym similarities based on the transformations learned during the training. For example, the vector representations of `Manchtftcr' and `Manchester' are more similar (i.e., the vectors are close to each other) when they have been generated from a DeezyMatch model that has been trained on a dataset which encloses these types of transformations (in this case, OCR-induced). In this step, we compute the distance of each query vector with respect to all gazetteer vectors and rank them according to the distance. We use \textit{faiss}, a library for efficient similarity search, to compute the $L_{2}$-norm distances \cite{Johnson_2019}.  
    \end{enumerate}
    
    In practice, the gazetteer usually has many more entries than the number of queries (i.e.~toponyms for which we want to find candidates). As remarked above, an advantage of the proposed method is that vector representations for the gazetteer are computed only once (for a given trained model). For all subsequent queries, only the query vectors are generated and compared to the gazetteer vectors. This significantly reduces the computation time compared to more traditional string-matching methods (e.g.~Levenshtein distance) in which one query is compared to \emph{n} possible variations of all potential candidates in each run. DeezyMatch also supports on-the-fly ranking, that is, the toponym queries are converted into vector representations and compared to the gazetteer vectors automatically.
    
\section{Data and Resources}\label{sec:data_resources}
In this section we introduce the datasets and resources used in the experiments of Section \ref{sec:experiments}. We first describe the datasets that we use for the downstream task of selecting candidates for toponyms in text, in Section \ref{subsec:candselect_datasets}. They inform the choice of the gazetteers that we use both for creating the datasets for \textit{toponym matching} and evaluating the performance of \textit{candidate selection}. The gazetteers used in our experiments are described in Section \ref{subsec:gazetteers}. In Section \ref{subsec:topmatch_datasets}, we describe the datasets used for training and evaluating the toponym matching models.

    \subsection{Candidate Selection Datasets}\label{subsec:candselect_datasets}
    To assess the performance of geographical candidate selection, we use two datasets in English and one in Spanish. They are historical datasets (from the nineteenth and seventeenth centuries), and have been selected because they present interesting challenges, including diachronic or spelling variations and OCR errors. The three datasets consist of text documents in which toponyms have been recognized and resolved to latitude and longitude coordinate points. We discuss the characteristics of these datasets in the following paragraphs, and summarize them in Table \ref{tab:datasetsCS}.
    
    \subsubsection{War of the Rebellion corpus \textbf{(WOTR)} \cite{delozier2016creating}}
    This corpus is composed of historical texts in English (historical letters and reports) from the American Civil War (1861-1865), manually annotated with geographic references. The documents had been previously OCR'd and manually-corrected. Annotators were free to obtain coordinates from different sources. However, they were specifically shown how to retrieve them from Wikipedia pages, which therefore was the most-used resource. Larger geographical entities in the WOTR dataset (e.g. countries or states) were annotated both with geographical points and polygons. For consistency with the other datasets, we only considered points. We use their test set for our experiments, which has 1,479 annotated toponyms (of which 584 are unique, after lower-casing).
    
    \subsubsection{British Newspaper Archive toponym corpus \textbf{(BNA-FMP)}}
    The second dataset has been created as part of our project\footnote{Living with Machines: \url{http://livingwithmachines.ac.uk/}.} from historical newspaper articles in English obtained from the British Newspaper Archive, abbreviated \textbf{\textit{BNA-FMP}}.\footnote{Newspaper data was kindly shared by Findmypast: \url{https://www.findmypast.co.uk/}.} This dataset consists of 1,248 toponyms (of which 509 unique toponyms, after lower-casing) from 191 articles published between 1780 and 1870 in local newspapers based in Manchester and Ashton-under-Lyne (broadly representing the industrial north of England), and Dorchester and Poole (representing the rural south). We selected articles that are between 150 and 550 words long and with an OCR confidence score greater than 0.7, as reported in the metadata. We did not correct errors produced in the OCR or layout recognition steps. The annotator was asked to recognize every location mentioned in the text and map it to the URL of the Wikipedia article that refers to it. We then derived the latitude and longitude of the entry in question from WikiGazetteer, a Wikipedia-based gazetteer enhanced with information from Geonames \cite{ardanuy2019resolving}. The toponyms recognized in this dataset often contain OCR errors (e.g.~`iHancfjrcter' for `Manchester', or `WEYBIOIJTII' for `WEYMOUTH'), spelling variations (e.g.~`Leipsic' for `Leipzig', or `Montpelier' for `Montpellier'), historical anglicizations and other form of foreign toponym domestications (e.g.~ `Kingstown' for `D\'{u}n Laoghaire', `Queenstown' for `Cobh', or `Carlowitz' for `Sremski Karlovci'), name changes due to external factors (e.g.~`Constantinople' for `Istanbul'), or a combination of these. Out of the 509 unique toponyms in the dataset, there are 167 toponyms for which a true referent exists in the gazetteer but cannot be directly retrieved because there is no exact-matching toponym in the gazetteer. In most cases, this is due to OCR errors (111 instances) or spelling variations (26 instances).
    
    \subsubsection{La Argentina Manuscrita \textbf{(ArgManuscrita)}} \label{sss:argman} This dataset has been created as part of the Digital Humanities project \textit{La Argentina Manuscrita},\footnote{A description of the project is found here: \url{https://arounddh.org/en/la-argentina-manuscrita}, the data is openly available here: \url{https://recogito.pelagios.org/document/wzqxhk0h3vpikm}.} which used the semantic annotation tool Recogito\footnote{Recogito, an initiative of Pelagios Commons, \url{http://recogito.pelagios.org/}.} to geolocate toponyms from a seventeenth-century chronicle and travelogue in Spanish, describing the area around the Río de la Plata basin. This dataset is in Spanish and is composed of 799 toponyms (of which 200 unique toponyms, after lower-casing), and has been annotated with coordinates from different gazetteers, such as Geonames, Pleiades,\footnote{\url{http://pleiades.stoa.org/}.} and HGIS de las Indias.\footnote{\url{https://www.hgis-indias.net/}.}
    
    \begin{table}[h!]
        \begin{tabular}{lccc}
            \toprule
            Dataset & Unique toponyms & Language & Period \\
            \midrule
            ArgManuscrita & 200 & Spanish & 1610s \\
            WOTR (test) & 584 & English & 1860s \\
            BNA-FMP & 509 & English & 1780-1870 \\
            \bottomrule
        \end{tabular}
        \caption{Candidate selection datasets}
        \label{tab:datasetsCS}
    \end{table}

    \subsection{Gazetteers}\label{subsec:gazetteers}
    Gazetteers are geographical dictionaries. They can be either global (aiming at worldwide geographic coverage) or local (covering a specific region or time period). The choice of the gazetteer(s) to which toponyms are linked inevitably has an impact on candidate selection and therefore final resolution. For many digital humanities applications, a gazetteer for toponym resolution should faithfully reflect the geographical knowledge of the writer and intended audience of the texts. Noisy gazetteers --- those containing anachronistic records --- not only complicate candidate selection, they also introduce geographical information that may have been unknown to people in a particular historical context. For example, if the goal is to resolve the toponyms in an English-translated collection of texts from second century Greece, it would be preferable that the city of Athens in Georgia is not even present in the gazetteer, so that it would not be retrieved as a possible candidate. Our candidate selection method is flexible in the choice of the gazetteer, as long as its entries (i.e.~places) include at least latitude and longitude coordinate points and potential alternate names that may be used to refer to them.
    
    Gazetteers serve two different goals in this paper: (1) to create sets of positive and negative pairs used in the toponym matching step to train the classifiers (toponym pair datasets are described in the following section, \ref{subsec:topmatch_datasets}), and (2) as the knowledge base against which we perform geographical candidate selection. We have created a WikiGazetteer \cite{ardanuy2019resolving} (a Wikipedia-based gazetteer enriched with Geonames data) for the different languages of the datasets described in Section \ref{subsec:candselect_datasets}, i.e.~English and Spanish, and a WikiGazetter for Greek, to test DeezyMatch in an alphabet different than Latin. The three gazetteers have been created from their corresponding (at the time) latest versions of Wikipedia.\footnote{See the instructions to create a WikiGazetteer from a specific Wikipedia version here: \url{https://github.com/Living-with-machines/lwm_GIR19_resolving_places/tree/master/gazetteer_construction}.} These three WikiGazetteers have been used to create the sets of positive and negative pairs that will be described in the following section. Besides, for the candidate selection experiments we have expanded the Spanish WikiGazetteer with the \textit{HGIS de las Indias} gazetteer,\footnote{To learn more, see \citet{stangl2018empire} and the webpage: \url{https://hgis.club/historical-geography-of-bourbon-spanish-america}.} which collects the historical geography of colonial Spanish America and corresponds with the period of the \textit{Argentina Manuscrita} dataset (described in section \ref{sss:argman}).
    The four resulting gazetteers are summarized in Table \ref{tab:summary_gazetteers}.
    
    \begin{table}[h!]
      \begin{tabular}{lccc}
        \toprule
        Gazetteer & Language & Locations & Unique altnames \\
        \midrule
        WG:en\_gz & English & 1,144,016 & 2,455,966 \\
        WG:es\_gz & Spanish & 338,239 & 550,697 \\
        WG:es+HGISIndias & Spanish & 351,040 & 556,985 \\
        WG:el\_gz & Greek & 21,037 & 34,572 \\
      \bottomrule
    \end{tabular}
      \caption{Summary of gazetteers: \textit{Language} indicates the language of the Wikipedia version from which the gazetteer has been built, \textit{Locations} is the number of entity entries in the gazetteer, and \textit{Unique altnames} is the number of unique place names present in the gazetteer.}
      \label{tab:summary_gazetteers}
    \end{table}
    
    \subsection{Toponym Matching Datasets}\label{subsec:topmatch_datasets}
    We use several datasets to evaluate the intrinsic performance of our toponym matching approach: an existing dataset, introduced in \citet{santos2018learning} (Section \ref{subsub:santos2018}), and two new realistic datasets that cover different types of variations (Sections \ref{subsub:wikigaz} and \ref{subsub:19thCdataset}). They are summarized in Table \ref{tab:datasetsTM}.
    
    \subsubsection{\citet{santos2018learning} (\textbf{Santos})}\label{subsub:santos2018}
    GeoNames is a large gazetteer that is publicly available. Each location is associated with multiple names (i.e.~corresponding to historical or regional denominations, or to names in different languages and alphabets, etc.). \citet{santos2018learning} generated a dataset of 5M toponym pairs from Geonames alternate names, half of which are matching pairs. A matching pair of toponyms consists of two alternate names that correspond to the same entity (e.g.~`London' and `Londres'), as long as both names are longer than two characters and they are not identical after lower-casing. A non-matching pair consists of alternate names that do not correspond to the same entity (e.g.~`Salsipuedes' and `Isla San Pedro'). To make sure not all non-matching pairs are completely dissimilar, the authors discarded pairs with a Jaccard similarity equal to zero with a probability of 0.75. This resource is the largest employed in our work and contains toponym pairs from different languages and alphabets.
    
    \subsubsection{WikiGazetteer datasets}\label{subsub:wikigaz}
    We created the Spanish, English, and Greek versions of WikiGazetteer to build three new datasets of toponym pairs, created in a similar way as in \citet{santos2018learning}. The resulting datasets (\textbf{WG:es}, \textbf{WG:en}, and \textbf{WG:el}) are significantly smaller, less ambitious (they do not have toponym pairs from across alphabets) and are biased towards place names in Spanish, English, and Greek respectively, which fitted our downstream scenarios better. Half of the dataset is composed by trivial cases: negative pairs that are extremely dissimilar (e.g.~`London' and `Paris') and positive pairs that are either identical or nearly-identical (except for differences in letter case). We found trivial negative cases by randomly selecting 50 toponyms from the gazetteer, sorting them from more dissimilar to less to the source toponym, and selecting the top most dissimilar. The other half of the dataset is comprised of very challenging cases, to force the model to learn the more nuanced toponymic variations (e.g.~`Edinburgh' and `Edinborg' as positive, and `Sheverin' and `Neverin' as negative). To do so, we considered as positive matches alternate names that can correspond to the same entity and that have a normalized Levenshtein-Damerau similarity of above 0.25. We then created non-matching pairs by collecting the most similar alternate names for a toponym and ranked them using normalized Levenshtein-Damerau distance, removing alternate names of entities that are within a distance of 50 km\footnote{We added distance as a restriction to minimize the incidence of highly-related, though still distinct, entities in the gazetteer, such as `Barcelona' and `Barcelon\`{e}s' (its enclosing administrative territorial entity) or `Port of Barcelona' (a nested entity).} of each other. We chose this threshold heuristically: it is conservative enough to filter out unwelcome noise, while at the same time allows finding enough potential non-matching pairs. We made sure that for each toponym there were as many positive as negative pairs. This resulted in a balanced dataset for each WikiGazetteer (see examples of trivial and challenging positive and negative pairs for toponym `Aintourine' extracted from the English WikiGazetteer dataset \textit{WG:en\_gz} in table \ref{tab:wgendataset}.).
    
    \begin{table}[h!]
        \begin{tabular}{llc}
            \toprule
            Toponym 1 & Toponym 2 & Matching  \\
            \midrule
                Aintourine & Aintourine & True  \\
                Aintourine & AINTOURINE & True  \\
                Aintourine & Haagsche Bosch & False  \\
                Aintourine & Sorkhankalateh & False  \\
                Aintourine & Am Toûrîne & True  \\
                Aintourine & Aïn Toûrîne & True  \\
                Aintourine & Tigantourine & False  \\
                Aintourine & Tiguentourine & False  \\
            \bottomrule
        \end{tabular}
        \caption{Positive and negative toponym pairs extracted from \textit{WG:en\_gz}.}
        \label{tab:wgendataset}
    \end{table}
    
    \subsubsection{Nineteenth-century English OCR dataset (\textbf{OCR})}\label{subsub:19thCdataset}
    \citet{evershed2014correcting} released a corpus of OCR'd newspaper texts\footnote{The data comes from the National Library of Australia Trove digitized newspaper collection.} that were aligned with corrections performed by volunteers. Following the procedure described in \citet{van2020assessing}, we aligned the texts at the token level and identified tokens recognized as being part of named entities in the human-corrected text. Since the goal of this dataset is to allow learning OCR transformations, we filtered out pairs of aligned tokens if: (1) the OCR'd and its correction are identical, (2) the OCR'd token has less than two characters, (3) the OCR'd text is exactly a substring of the human-corrected text, or vice-versa, (4) the human correction does not contain a hyphen,\footnote{We decided to filter out tokens containing hyphens because of hyphenated words at the end of the line, sometimes resulting in partial tokens matching full tokens.} (5) the correction is composed only of alphabetical characters, and (6) the edit operation that transforms one token into another (e.g.~`c' into `e' in the pair `Jagclman-Jagelman') occurs more than once in the dataset. We then created a dataset that has similar characteristics to the \textit{Santos} and WikiGazetteer-based datasets: for each human-corrected token, we consider all its observed OCR'd variations in the dataset as positive pairings. We then capture the most observed OCR transformations in the dataset, and artificially build negative pairs by introducing unobserved random transformations for characters in the human-corrected string. We build as many negative pairs as positive pairs exist for a corrected string. See some examples in Table \ref{tab:ocrdataset}.
    
    \begin{table}[h!]
        \begin{tabular}{llc}
            \toprule
            Correction & Variation & Matching  \\
            \midrule
            Zurich     & Zmich     & True \\
            Zurich     & 7urich    & True \\
            Zurich     & Zuiich    & True \\
            Zurich     & Zunch     & True \\
            Zurich     & Zururn    & False \\
            Zurich     & Zur»/     & False \\
            Zurich     & ZuhSch    & False \\
            Zurich     & Zuâch     & False \\
            \bottomrule
        \end{tabular}
        \caption{Positive and negative OCR pairs}
        \label{tab:ocrdataset}
    \end{table}
        
    \subsubsection{Summary of datasets}
    Table \ref{tab:datasetsTM} summarizes the key characteristics of the different datasets that we have used to create the models and assess the performance of our toponym matching component. As a post-processing step, for each of the five datasets we removed pairs if one of the elements was an empty string, and removed duplicates (including reverse duplicates such as `Florence, Firenze, True' and `Firenze, Florence, True'). We removed, for each true pair, a corresponding false pair, and vice versa. Finally, for each dataset, we provide a balanced training/validation and a test set (90\% and 10\% of the whole dataset respectively).
    
    \begin{table}[h!]
    \begin{tabular}{lccc}
        \toprule
        Dataset & Toponym Pairs & Source & Alphabet \\
        \midrule
        Santos \cite{santos2018learning} & 4,337,446 & Geonames & Multiple \\
        WG:en & 669,376 & Wikigaz (EN) & Latin \\
        WG:es & 152,026 & Wikigaz (ES) & Latin \\
        WG:el & 3,086 & Wikigaz (EL) & Greek \\
        OCR & 93,111 & OCR & Latin \\
        \bottomrule
    \end{tabular}
    \caption{Toponym matching datasets}
    \label{tab:datasetsTM}
\end{table}
    
\section{Experiments}\label{sec:experiments}
Our approach is built around two main components, toponym matching and candidate selection. We assess the performance of each of them in the next two sub-sections.

    \subsection{Toponym Matching}
    The goal of toponym matching is to assess whether two strings can refer to the same location. In this section, we evaluate the performance of our toponym matching component in different settings and scenarios and in comparison with well-established baselines.
    
        \begin{figure*}[h]
          \includegraphics[width=\textwidth]{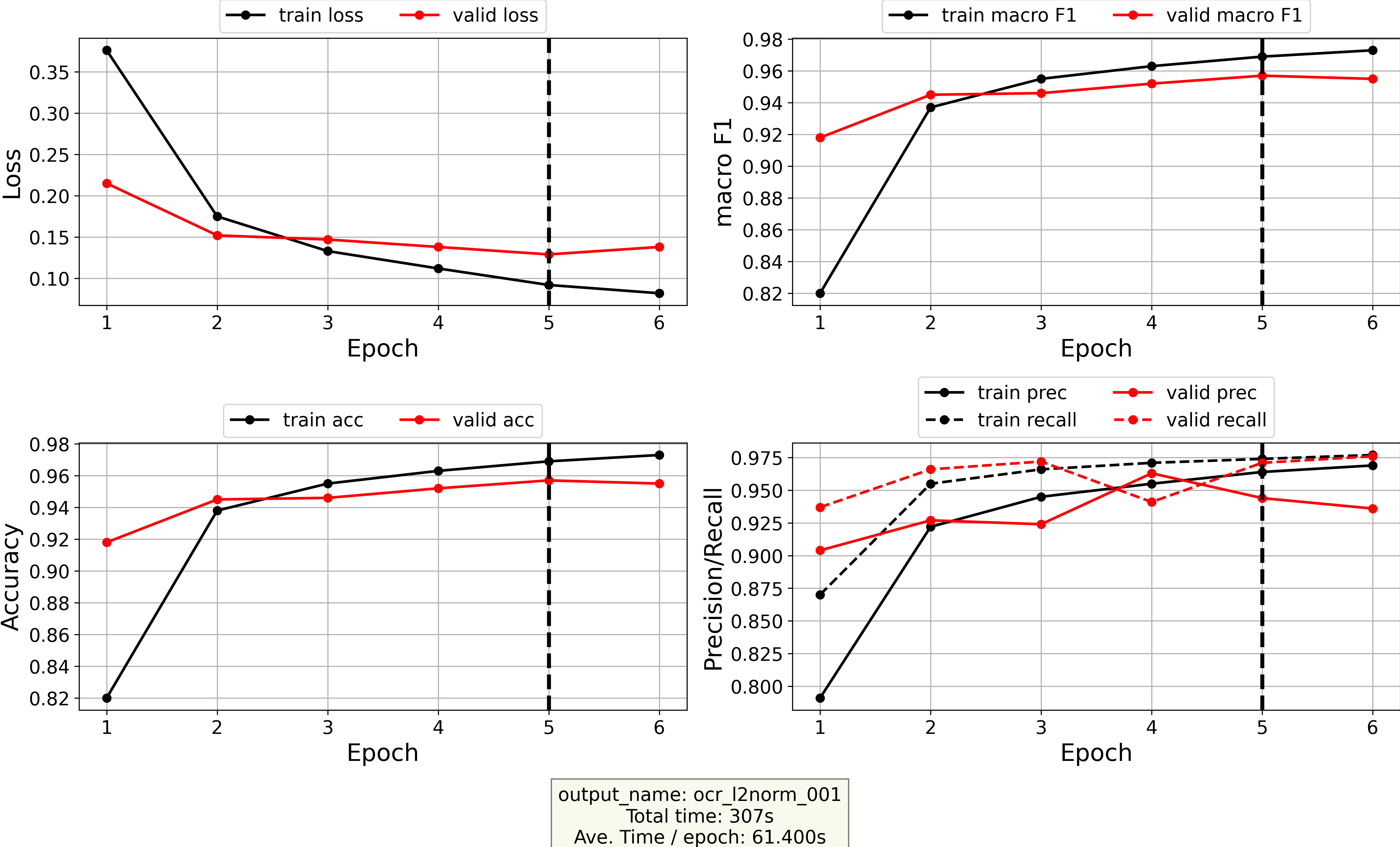}
          \caption{Tracking metrics during a model training. DeezyMatch logs various evaluation metrics at each epoch: (a) train and validation losses; (b) macro F1 scores (the harmonic mean of the precision and recall); (c) Accuracy on both train and validation sets; (d) precision and recall. The selected model (epoch 5 in this example) is shown by vertical dashed lines.}
        \label{log_ocr}
        \end{figure*}
        
        \subsubsection{DeezyMatch Models}
        \label{subsec:deezy_models}
        The DeezyMatch models used in this study have similar neural network architectures and hyperparameters. In all models, the underlying dataset is preprocessed by normalizing the text to the ASCII encoding standard, by removing both the leading and the trailing empty characters, and by adding a prefix and suffix (character `|') to the string. We keep the letter case in toponym pairs. The training/validation datasets are used for training, hyperparameter tuning and model selection. The test set is used for reporting the final results. A character-level embedding is employed to convert the preprocessed text into vectors of size 60. The two embedding vectors of a toponym pair are then fed (in batches of size 64) to two parallel bi-directional GRUs with two layers. Each GRU network has a hidden state of size 60 and a maximum sequence length of 120. The learnable parameters (i.e.~weights and biases) of the two GRUs are shared, which helps the model to learn transformations regardless of the order of toponyms in an input pair. Each bi-directional GRU network outputs two vectors corresponding to the last hidden states of the forward and backward passes. These two vectors are then concatenated to form one vector with the length of 2$\times$hidden-state-size (i.e., 120) per GRU. We call them $h_{GRU1}$ and $h_{GRU2}$ for the first and second networks, respectively. DeezyMatch supports different ways of combining these two vectors. In our experiments, we create one vector for each toponym pair: $1 - |h_{GRU1} - h_{GRU2}|^2$. The resulting vector is then passed to a feedforward neural network with one hidden layer of size 120 with ReLU activation functions and one output unit with sigmoid nonlinearity. We use the Binary Cross Entropy criterion and the Adam optimization method with a learning rate of 0.001 to adjust the learnable parameters in our model (591,122 parameters in total). A dropout probability of 0.01 was used in all layers (i.e., GRUs and fully-connected layers) for regularization. To avoid overfitting, we also use early stopping and select the model where the validation loss starts to increase.
        
        \subsubsection{Baselines}\label{sss:baselines}
        As a baseline, we use normalized Levenshtein-Damerau edit distance,\footnote{As done in previous work \cite{santos2018toponym}, we used the \texttt{pyxDamerauLevenshtein} python implementation: \url{https://pypi.org/project/pyxDamerauLevenshtein/}.} a traditional string similarity measure based on the number of operations needed to transform one string to another, to classify pairs of toponyms as either matching or not. We find the optimal threshold on the training/validation set. For comparison, we also report the performance of the toponym matching implementation by \citet{santos2018toponym}.\footnote{We had to slightly modify the original implementation to be compatible with the Tensorflow backend. We have notified the authors.}
    
        \subsubsection{Metrics}
        Following \citet{santos2018toponym}, we treat toponym matching as a binary classification task. We report the F-Score (i.e.~the harmonic mean between precision and recall).\footnote{The inference function of DeezyMatch additionally provides accuracy, precision and recall. We have not included them in the paper for readability reasons, but all experiments can be found in our Github repository: \url{https://github.com/Living-with-machines/LwM_SIGSPATIAL2020_ToponymMatching}. We provide all resources to allow full reproducibility of the results.} 
        
        \subsubsection{Evaluation}
        Table \ref{tab:topmatch_results} reports the  performance of DeezyMatch and comparing methods on several toponym matching datasets. Note that we tested the performance of \textit{DeezyMatch} and \textit{LevDam} on the test split (10\% of the datasets), to ensure reproducibility. For \citet{santos2018toponym}, this was not possible without extensively changing their code, as their implementation does not allow testing an existing model on new data. Therefore, we evaluated it only through two-fold cross validation on the training/validation dataset. Though not strictly comparable, we show in Table \ref{tab:topmatch_results} the differences in performance between the two implementations.\footnote{Performance of \citet{santos2018toponym} on their dataset is considerably lower than that reported in their paper (0.89 F1 Score). This difference is due to the fact that we removed all duplicates (including reverse duplicates, such as `Yangji-mal, yangjimal' and `yangjimal, Yangji-mal') in the dataset ($\sim$7\% of the original resource). } 
        
        \begin{table}[h!]
          \begin{tabular}{lccccc}
            \toprule
            & Santos & WG:en & WG:es & WG:el & OCR \\
            \midrule
            LevDam & 0.70 & 0.74 & 0.75 & 0.83 & 0.76 \\
            \citet{santos2018toponym} & 0.82 & 0.92 & 0.90 & 0.80 & 0.95 \\
            \midrule
            DeezyMatch & 0.89 & 0.94 & 0.92 & 0.84 & 0.95 \\
          \bottomrule
        \end{tabular}
          \caption{Evaluation of Toponym Matching methods in terms of F1 Score on the different datasets described in Section \ref{subsec:topmatch_datasets}.}
          \label{tab:topmatch_results}
        \end{table}
        
    \subsection{Candidate Selection}
    \label{subsec:candidate_selection}
    Candidate selection is the task of ensuring that the correct entity is found among the retrieved candidates. In this section, we report the performance of our candidate selection component on different toponym resolution datasets.
    
        \subsubsection{Metrics}\label{sec:candrank_metrics}
        Evaluation of geographical candidate selection in toponym resolution systems is not always straightforward. This is in part due to the lack of a true gold standard for places, as gazetteers indicate the position of a place on the Earth's surface through its approximate coordinates, which may not coincide with the same exact coordinates used by the dataset annotators. Because of this, it has been common in the literature to allow an error distance, be it in km (usually 161km, i.e.~100 miles) or degrees \cite{delozier2015gazetteer,roller2012supervised,speriosu2013text,cheng2010you}. We decided to be more restrictive and considered a candidate as correct if it was within 10km of its location in the gazetteer.\footnote{We observed that 161km was too large a distance for some of our datasets. A smaller window ensures higher reliability of our precision metrics.} Our candidate selection module finds potential matching toponyms in the gazetteer, not entities. 
        During evaluation, if a toponym in the gazetteer can refer to more than one entity, we select the one closest to the gold standard coordinates; and if this falls within 10km from the gold standard location, we consider it a true positive.
        
        Candidate selection is the task of ensuring that the correct entity is found among the retrieved candidates. 
        We report our results based on two metrics: precision at 1 candidate (\textit{P@1}) and mean average precision at 5, 10, and 20 candidates (\textit{MAP@5}, \textit{MAP@10}, and \textit{MAP@20}), to evaluate the quality of the ranking.\footnote{Given the lack of a true gold standard mentioned above, some true candidates are incorrectly considered as false positives. Chile is an extreme example of this: the annotator assigned coordinates with -37.78 latitude and -71.36 longitude to this country and the coordinates in the \textit{WG:es+HGISIndias} gazetteer are -33.45 latitude and -70.67 longitude, almost 500km apart; and yet, they are both correctly a point in Chile. In our evaluation results, we excluded cases where none of the methods retrieved any correct results, because our intention is not to evaluate gazetteer-to-dataset compatibility, but the quality of our method's candidate selection compared to other methods.}
        
        \subsubsection{Baselines}
        To better understand the performance of DeezyMatch for candidate selection, we provide the following baselines: (1) \textit{Exact}: a candidate is retrieved if it exactly matches the toponym in the text (case insensitive), which is the most common type of candidate selection in entity linking and toponym resolution methods; and (2) \textit{LevDam}: candidates are ranked by string similarity, based on normalized Levenshtein-Damerau edit distance.\footnote{We used the \texttt{pyxDamerauLevenshtein} implementation: \url{https://pypi.org/project/pyxDamerauLevenshtein/}.} While this is a strong baseline, it is often impracticable in downstream toponym resolution because of time complexity (see Table \ref{tab:candranker_eval}).
        
        \begin{table*}[h!]
        \centering
          \begin{tabular}{lccccccc}
            \toprule
            & Gazetteer & P@1 & MAP@5 & MAP@10 & MAP@20 & Time \\
            \midrule
            ArgManuscrita:exact & WG:es\_gz+HGISIndias & 0.69 & - & - & - & -  \\
            ArgManuscrita:LevDam & WG:es\_gz+HGISIndias & 0.78 & 0.77 & 0.72 & 0.70 & 29.77m  \\
            ArgManuscrita:DeezyMatch (WG:es) & WG:es\_gz+HGISIndias & 0.78 & 0.78 & 0.76 & 0.74 & 0.73m \\
            \midrule
            Wotr:exact & WG:en\_gz & 0.86 & - & - & - & - \\
            Wotr:LevDam & WG:en\_gz & 0.92 & 0.89 & 0.84 & 0.80 & 308m \\
            Wotr:DeezyMatch (WG:en) & WG:en\_gz & 0.93 & 0.92 & 0.90 & 0.87 & 4.75m \\
            \midrule
            BNA-FMP:exact & WG:en\_gz & 0.77 & - & - & - & - \\
            BNA-FMP:LevDam & WG:en\_gz & 0.92 & 0.88 & 0.82 & 0.76 & 120m \\
            BNA-FMP:DeezyMatch (WG:en) & WG:en\_gz & 0.85 & 0.85 & 0.82 & 0.78 & 4.27m \\
            BNA-FMP:DeezyMatch (OCR) & WG:en\_gz & 0.83 & 0.83 & 0.82 & 0.80 & 4.27m \\
          \bottomrule
        \end{tabular}
          \caption{DeezyMatch candidate ranker performance. All DeezyMatch models have been trained using the model architectures and the choice of hyperparameters described in Section \ref{subsec:deezy_models}. The datasets on which they have been trained are specified in parentheses in the first column; \textit{Gazetteer} specifies the gazetteer from where candidates are retrieved for each scenario. All methods are evaluated using the same metrics (columns \textit{P@1}, \textit{MAP@5}, \textit{MAP@10}, and \textit{MAP@20}). \textit{Time} indicates total computation time on CPU, which mostly depends on the number of queries and the size of gazetteer.}
          \label{tab:candranker_eval}
        \end{table*}
        
        \subsubsection{Evaluation and Discussion}
        We report the performance of our method and baselines on three datasets for candidate selection in Table \ref{tab:candranker_eval} (datasets and gazetteers are described in detail in Sections \ref{subsec:candselect_datasets} and \ref{subsec:gazetteers}, respectively). We show how the \textit{exact} baseline, while being the most common approach in entity linking systems, is insufficient for the task of toponym candidate selection from noisy datasets. We observe that the performance of the \textit{LevDam} and \textit{DeezyMatch} methods varies significantly depending on the dataset: while DeezyMatch clearly outperforms LevDam on the \textit{Wotr} dataset and on the \textit{ArgManuscrita} dataset (in the second case, in terms of MAP@5, MAP@10, and MAP@20, while being comparable in P@1), our method shows lower performance on the BNA-FMP dataset in comparison with the LevDam baseline, particularly in terms of P@1.
        
        We argue that the reason behind these differences in performance is found either in the nature of the toponymic variations present in the datasets or in the nature of the toponym matching datasets, from which we learned the transformations. To better understand this, for each dataset we investigated the results by looking at the selected candidates correctly retrieved by one method and failed to be retrieved by another. DeezyMatch seems clearly better than LevDam when the transformation affects significant part of the toponym; this is particularly the case of long multi-token place names. In the \textit{ArgManuscrita} dataset, for example, given the toponym `provincia del Paraguay', DeezyMatch returns `Republica del Paraguay', `República del Paraguay', `Paraguai - Paraguay', `Republic of Paraguay', and `Departamento Alto Paraguay' as most likely candidates, whereas LevDam returns `Provincia de Paragua', `Provincia de Veragua', `Provincia de Camaguey', followed by a large number of other provinces from around the world, sorted by string surface similarity. Just to provide another example, DeezyMatch ranks `Departamento de Tarija' as the most likely match for toponym `corregimiento de Tarija', while LevDam retrieves `corregimiento de Tunja', followed by other `corregimientos' (a type of country subdivision), such as `corregimiento de Loja'. DeezyMatch is able to rank these candidates better because it has learned similar transformations from the corresponding toponym matching dataset (in this case, \textit{WG:es}), which, even though it does not have the alternate names `corregimiento de Tarija' and `provincia del Paraguay' in it, has other similarly-shaped multi-token toponyms where it can learn these transformations from.
        
        On the contrary, LevDam is a very strong baseline when the string surface difference between the two toponyms is very small (i.e.~very few characters difference). This is particularly common in the \textit{BNA-FMP} dataset, where most toponym variations are caused by OCR errors. While \textit{DeezyMatch} offers better results in comparison with \textit{exact} matching, and while the quality of its ranking remains constantly high when more candidates are retrieved, it is clearly behind \textit{LevDam} in retrieving the best first candidate (P@1). In order to better understand this issue, we tried two \textit{DeezyMatch} models, one trained on \textit{WG:en} and one trained on \textit{OCR}. However, disappointingly, the second model produced worse results. This might be due to the fact that, while some typical OCR transformations seem to be correctly enclosed in the OCR-based DeezyMatch model, they are not always aligned between the toponym matching dataset and the OCR errors in the gold standard candidate selection dataset. The \textit{OCR} matching dataset does not in fact have the same origin as the \textit{BNA-FMP} dataset, and while some OCR transformations are probably largely generic (e.g.~`e' to `c', or `B' to `P'), different typographies or OCR softwares may lead to learning unwelcome transformations. On the other hand, by using a model trained only on the OCR dataset, we are disregarding the other types of transformations that are enclosed in more generic toponym-based resources, such as \textit{WG:en}. We will continue our experiments on OCR-induced noise in future publications, considering other hyperparameters and exploring transfer learning approaches, a functionality that \textit{DeezyMatch} already provides.
        
        Nevertheless, whereas \textit{LevDam} is generally a strong baseline, its high computational cost makes it impracticable to use in many real applications of candidate selection. In this regard, \textit{DeezyMatch} undeniably offers a strong alternative. While training the model and generating the gazetteer candidate vectors is computationally expensive,\footnote{DeezyMatch training time (on GPU) until validation loss starts to increase: Santos (4,337,446 pairs): 10h4, WG:en (669,376): 56m, WG:es (152,026): 21m, WG:el (3,086): 1m, OCR (93,111): 5m. Generating candidate vectors for the largest gazetteer (i.e.~WG:en\_gz, with 2,455,966 unique alternate names) takes 204m on CPU.} these are steps that need to be done only once and can be reused for all following candidate selection tasks that employ the same gazetteer.\footnote{Query vector representations can be generated, compared to candidate vectors, and ranked on-the-fly. This can be directly integrated into an entity linking pipeline.} Time needed for generating query vectors given a set of toponyms and finding candidates in a gazetteer is reported in the last column of Table \ref{tab:candranker_eval} and is significantly lower than \textit{LevDam} performance in all cases.

\section{Conclusion}

In this paper, we discussed the importance of precisely identifying candidates in order to resolve toponyms to their real-world referents. In particular, we highlighted its necessity when working with noisy and non-standard texts (e.g.~documents digitized with OCR). To foster further research on this intermediary step, we have introduced DeezyMatch, a flexible deep learning method for candidate selection through toponym matching. It is based on the state-of-the-art neural network architectures and has been tested in different evaluation settings, considering various challenging scenarios (cross-lingual, diachronic, and regional variations, as well as OCR errors) and in comparison with a series of well established baselines. DeezyMatch, the evaluation framework presented in this paper, and all other resources employed are useful contributions to other researchers working at the intersection of geospatial information retrieval and digital humanities.

\bibliographystyle{ACM-Reference-Format}
\bibliography{sample-base}

\clearpage

\end{document}